\documentclass[10pt,twocolumn,letterpaper]{article}

\usepackage[pagenumbers]{cvpr} 

\definecolor{cvprblue}{rgb}{0.21,0.49,0.74}
\usepackage[pagebackref,breaklinks,colorlinks,allcolors=cvprblue]{hyperref}

\def\paperID{***} 
\def\confName{CVPR}
\def\confYear{2025}

\usepackage{hyperref}
\usepackage{graphicx}
\usepackage{multirow}

\usepackage{algorithm}
\usepackage{alphalph}
\usepackage{algorithmic}
\usepackage{bbding}
\usepackage{makecell}
\usepackage[accsupp]{axessibility}

\usepackage{orcidlink}

\usepackage{enumitem}

\usepackage[utf8]{inputenc} 
\usepackage[T1]{fontenc}    
\usepackage{url}            
\usepackage{booktabs}       
\usepackage{amsfonts}       
\usepackage{nicefrac}       
\usepackage{microtype}      
\usepackage{amssymb}
\usepackage{pifont}

\newcommand{\name}{HyperSeg}
\newcommand{\dname}{Towards Universal Visual Segmentation\\with Large Language Model}

\title{\name: \dname}

\author{Cong Wei$^{1,2}$, Yujie Zhong$^{2}$$^{\dagger}$, Haoxian Tan$^{2}$, Yong Liu$^{1}$, Zheng Zhao$^2$, Jie Hu$^2$, and Yujiu Yang$^{1}$$^{\dagger}$  \\
$^1$Tsinghua Shenzhen International Graduate School, Tsinghua University 
   $^2$Meituan Inc.\\
{\tt\small weic22@mails.tsinghua.edu.cn,} 
{\tt\small jaszhong@hotmail.com,} 
{\tt\small yang.yujiu@sz.tsinghua.edu.cn}
}

\begin{document}
\twocolumn[{%
\renewcommand\twocolumn[1][]{#1}%
\maketitle

\vspace{-5mm}
\centering
\includegraphics[width=\textwidth]{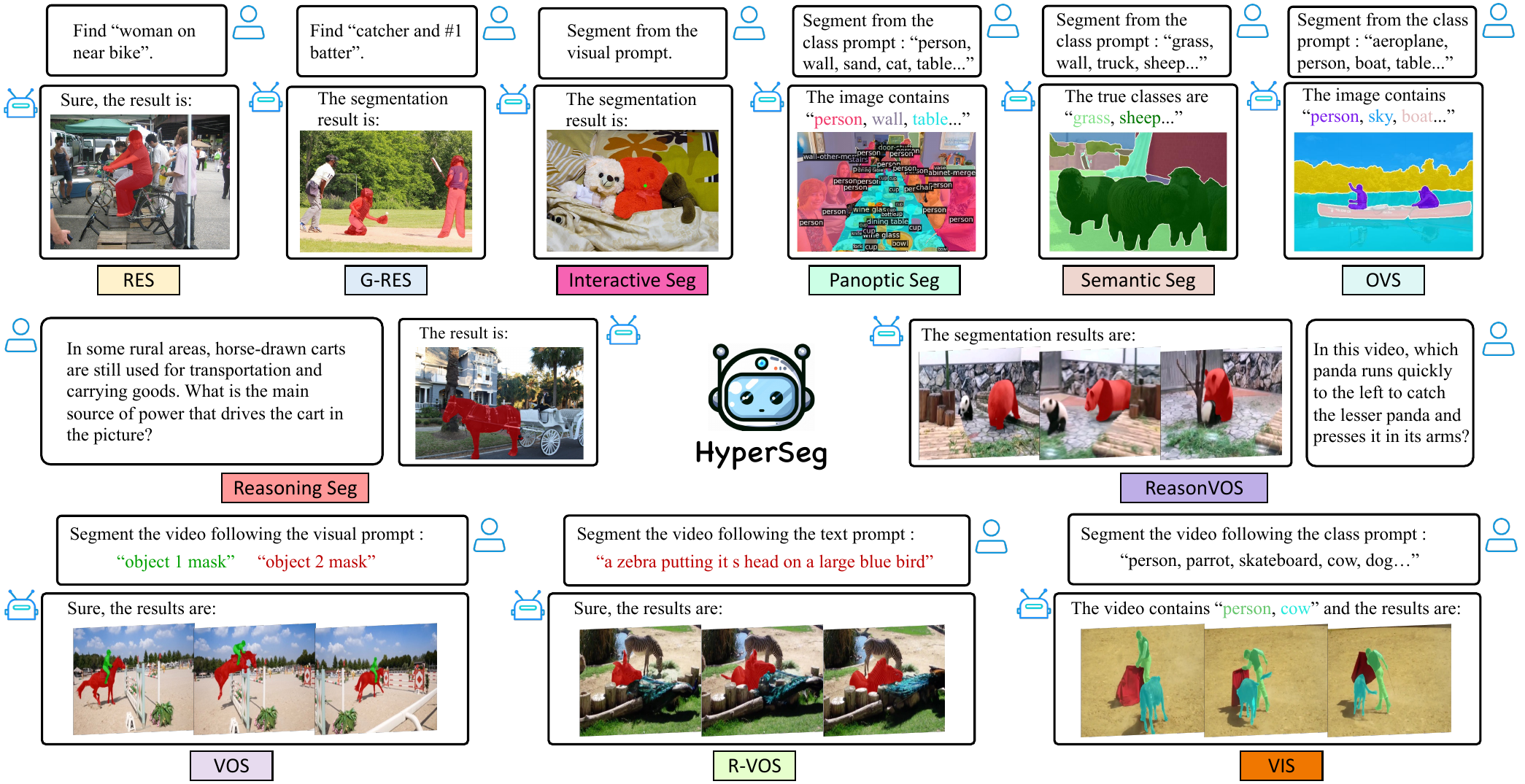}
\vspace{-2mm}
\captionof{figure}{Illustration of our \name~which can conduct image and video segmentation tasks with various language and visual instructions. Additionally, \name~can handle complicated reasoning perception tasks compared with previous universal segmentation methods. To our knowledge, \name~is the first VLLM-based universal segmentation model with perception and complex reasoning abilities in both image and video domains.}
\label{fig:intro}
\vspace{0.6cm}
}]


\renewcommand{\thefootnote}{\fnsymbol{footnote}}
\footnotetext{$^{\dagger}$Corresponding authors.}

\begin{abstract}
\indent This paper aims to address universal segmentation for image and video perception with the strong reasoning ability empowered by Visual Large Language Models (VLLMs). Despite significant progress in current unified segmentation methods, limitations in adaptation to both image and video scenarios, as well as the complex reasoning segmentation, make it difficult for them to handle various challenging instructions and achieve an accurate understanding of fine-grained vision-language correlations.
We propose HyperSeg, the first VLLM-based universal segmentation model for pixel-level image and video perception, encompassing generic segmentation tasks and more complex reasoning perception tasks requiring powerful reasoning abilities and world knowledge.
Besides, to fully leverage the recognition capabilities of VLLMs and the fine-grained visual information, HyperSeg incorporates hybrid entity recognition and fine-grained visual perceiver modules for various segmentation tasks. 
Combined with the temporal adapter, HyperSeg achieves a comprehensive understanding of temporal information. 
Experimental results validate the effectiveness of our insights in resolving universal image and video segmentation tasks, including the more complex reasoning perception tasks. Our code is available \href{https://github.com/congvvc/HyperSeg}{here}. 
\vspace{-4mm}

\end{abstract}

\section{Introduction}\label{sec:intro}
Visual segmentation is one of the most significant tasks in computer vision research, which aims to perform accurate pixel-level semantic understanding.
Many specialist models~\cite{he2017mask,cheng2022masked,jain2023oneformer,Lai2023LISARS} have made great progress in specific segmentation tasks while showing limitations in handling diverse and complicated scenarios since new training data, paradigms, and model architectures are required to adapt to new segmentation tasks. 
Recent works~\cite{lin2023uninext,zhang2023simple,li2024omg} propose a single framework to unify diverse segmentation tasks.
Despite promising, they show the inability to tackle text instructions and complex reasoning segmentation tasks needing powerful reasoning capabilities and world knowledge.

Visual Large Language Models (VLLMs) have exhibited excellent reasoning and conversation abilities, which play a pivotal role in various vision-language co-understanding tasks~\cite{cho2021unifying,liu2024visual,li2023blip,zhu2023minigpt,bai2023qwen}. 
\textbf{However, these methods are based on rudimentary vision-language alignment, which limits their ability to comprehend finer details in visual perception tasks, like pixel-level segmentation.}
Recent studies~\cite{Lai2023LISARS,wei2024lasagna,zhang2024psalm,yan2024visa,zhang2024omg} enables VLLMs to perform fine-grained visual understanding, like referring and reasoning segmentation. 
~\cite{Lai2023LISARS,wei2024lasagna,Ren2023PixelLMPR}
use the special token [SEG] generated by VLLMs as the prompt for the mask decoder to generate segmentation masks while ~\cite{zhang2024psalm,zhang2024omg} focus on incorporating instance-aware mask tokens into VLLMs.
\textbf{Though impressive, they show limitations to the universal segmentation framework based on VLLMs for both image and video domains and the capabilities of handling more complex video reasoning segmentation tasks.}

To this end, we introduce \name, the first VLLM-based universal segmentation model for pixel-level image and video perception with complex reasoning and conversation capabilities. \name~can conduct diverse image and video segmentation tasks with various
elaborate prompts and temporal adapter module. Besides, \name~shows excellent abilities in complicated vision-language reasoning perception tasks needing rich world knowledge, which is significant for real-world understanding and interactions.
As shown in Fig.~\ref{fig:intro}, the explored tasks contain both image and video domains.
We organize the tasks into two unified prompt formats: (1) text prompts (class names, reasoning questions, and referring languages), (2) visual prompts (box, mask, etc.). 
Owing to such flexible and cohesive design, \name~benefits from concurrent training on diverse segmentation tasks and vision domains, facilitating the intricate correlations between different instructions and visual concepts.
To further enhance fine-grained object perception and video understanding, we introduce three distinct designs in the following.

Firstly, we incorporate a hybrid entity recognition strategy to enhance the exploitation of VLLM's recognition capacity.
Generation-only works~\cite{Lai2023LISARS,xia2023gsva, Ren2023PixelLMPR} solely rely on VLLM for object prediction leading to poor performance in complex multi-object segmentation scenarios.
Decode-only methods~\cite{zhang2024psalm,zhang2024omg} use the prompt embedding and mask tokens decoded by VLLM to obtain class scores for each mask, which makes the mask tokens interact insufficiently with the semantic condition as they ignore the powerful generative capabilities of VLLM.
The proposed hybrid entity recognition leverages the VLLM's powerful generative abilities to enhance the mask tokens' comprehension of category semantics while maintaining the final class scores decoding process.

Secondly, previous VLLMs usually use coarse-level visual features obtained from CLIP~\cite{radford2021learning} series which primarily encode global visual information while overlooking visual details. 
To enhance VLLMs' ability of capturing visual details efficiently, we use the Fine-grained Visual Perceiver (FVP) to merge multi-scale visual features into fixed-length fine-grained tokens, allowing retrieval of rich visual details from various scales in the hierarchical vision encoder~\cite{cheng2022masked}. 

Thirdly,  recent VLLM-based segmentation methods~\cite{Lai2023LISARS,zhang2024psalm,zhang2024omg} demonstrate limitations in handling video perception tasks for video temporal understanding. To this end, 
we propose the temporal adapter for comprehensive video perception which incorporates global prompt aggregation and local space-time information injection for the coalescence of both long-term and short-term vision-language information.

Extensive experiments on various segmentation benchmarks demonstrate the preeminent segmentation ability of \name~, providing strong evidence of the effectiveness of our insights. Our \name~also exhibits promising performance on common Multi-modal benchmarks. Additionally, we explore the mutual influence among different tasks involving various visual and task types. 

Our contributions are summarized as follows: 
\begin{itemize}[leftmargin=0.5cm]
    \item We present \name, the first VLLM-based universal segmentation model for pixel-level image and video perception, covering a broad spectrum of common segmentation tasks, complex reasoning, and conversation-based vision-language understanding tasks.
    \item We incorporate hybrid entity recognition and fine-grained visual perceiver modules 
    to VLLM, which allow full exploitation of VLLM’s semantic recognition capacity and injection of fine-grained visual information to improve diverse detail-aware segmentation tasks.
    With the temporal adapter, \name~can conduct more challenging video perception tasks, achieving universal segmentation.
    \item \name~demonstrates superior capabilities on multiple segmentation tasks, achieving excellent performance on both generic and complex reasoning benchmarks with only one model.
    
\end{itemize}

\section{Related Work}
\textbf{Visual Large Language Model.}~
The emergence of Large Language Model (LLM) has significantly contributed to the development of VLMM. In this context, LLMs are enhanced with multimodal comprehension capabilities, allowing the vision-language co-understanding~\cite{li2023blip, alayrac2022flamingo, zhu2023minigpt, liu2024visual, liu2023improved, bai2023qwen}. 
Several notable examples of LLMs with multimodal comprehension include BLIP-2~\cite{li2023blip}, Flamingo~\cite{alayrac2022flamingo}, MiniGPT-4~\cite{zhu2023minigpt}, LLaVA~\cite{liu2024visual}, InstructBLIP~\cite{instructblip}, and Qwen-VL~\cite{bai2023qwen}. While these models have demonstrated impressive performance in vision-language tasks, they solely produce textual outputs that describe the entire image. This restricts their applicability in tasks that require the pixel-level detailed understanding.

\noindent \textbf{Perception with VLLM.}~
Several methods have been proposed to enhance VLLMs with a more detailed comprehension capability~\cite{chen2023shikra,wang2024visionllm,peng2023kosmos,you2023ferret,Lai2023LISARS,Ren2023PixelLMPR,rasheed2023glamm,pi2023perceptiongpt}.
Shikra~\cite{chen2023shikra}, Ferret~\cite{you2023ferret}, Kosmos-2~\cite{peng2023kosmos}, and VisionLLM~\cite{wang2024visionllm} are examples that provide grounding capabilities through regression of box coordinates. Conversely, LISA~\cite{Lai2023LISARS}, PixelLM~\cite{Ren2023PixelLMPR}, GLaMM~\cite{rasheed2023glamm}, and PerceptionGPT~\cite{pi2023perceptiongpt} employ a mask decoder to predict object masks from special tokens. 
Most of the existing methods utilize a next-token-prediction approach, which restricts their applicability. PSALM~\cite{zhang2024psalm} makes an important attempt to bring VLLM into visual perception tasks but fails to fully unleash the potential of VLLM. In contrast, our method propose to use a hybrid strategy to mitigate this problem and keep the capacity in high-level reasoning.

\noindent \textbf{Unified segmentation model.}~
Another line of studies focuses on the integration of various segmentation tasks into a single model. Mask2former~\cite{cheng2022masked} proposes a unified architecture that requires separate training on different segmentation tasks. 
OpenSeeD~\cite{zhang2023simple} introduces a text encoder and extends it to the Open-Set setting. Simultaneously, UNINEXT~\cite{lin2023uninext} supports referring segmentation with the assistance of text inputs and a text encoder. 
However, these works fall short of following complicated instructions and reasoning. In this work, we improve the understanding ability toward language by incorporating LLM, while also maintaining the original ability of vision-centric models. 
\section{Method} \label{sec:method}

\subsection{Overview}
\label{subsec:overview}

\begin{figure*}[t]
    \centering
    \includegraphics[width=\textwidth]{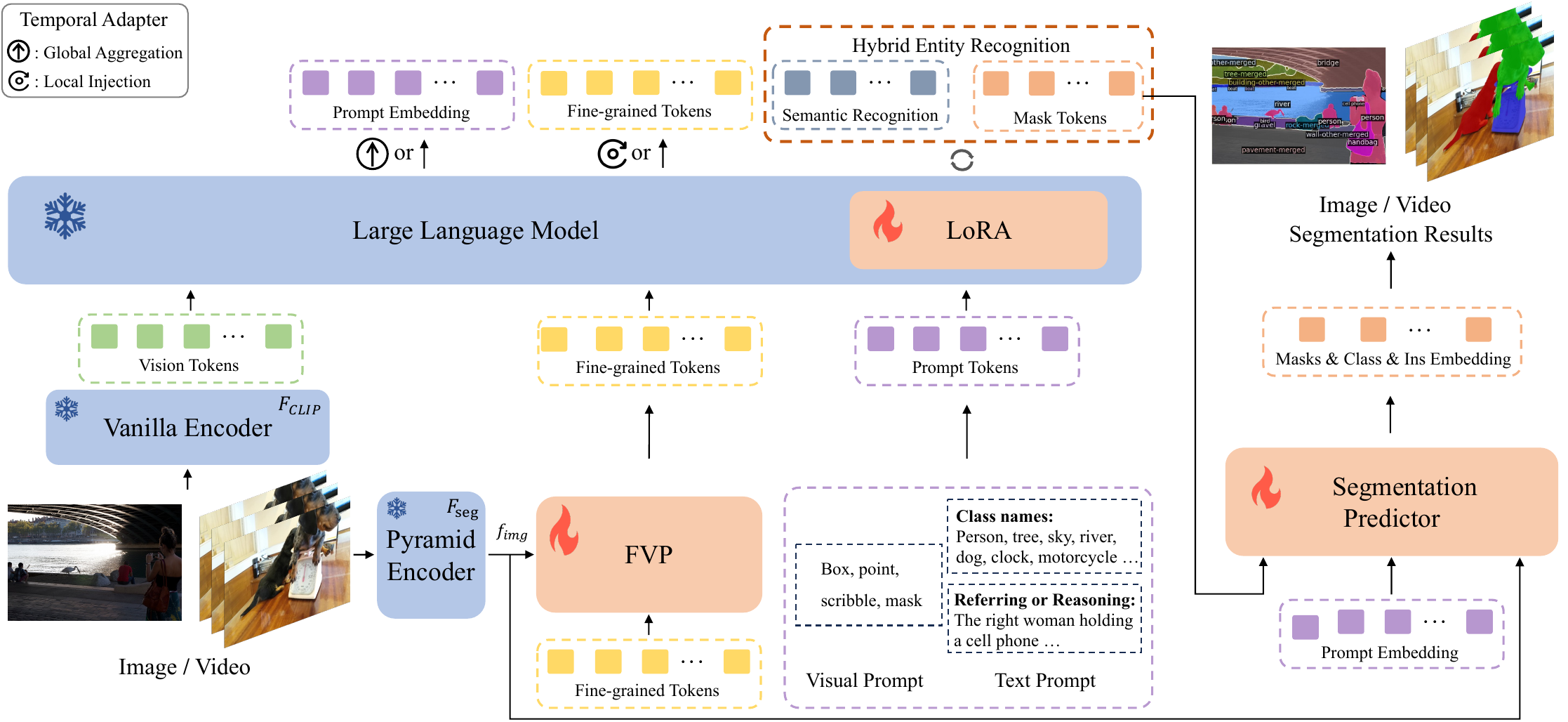}
    \caption{\textbf{Overview of \name}. \name~encodes the visual input in a multi-grained manner and concatenates the prompt for different perception tasks. We feed learnable fine-grained tokens into a Fine-grained Visual Perceiver (FVP) to integrate multi-scale high-resolution image features into LLM for detailed visual learning and to facilitate space-time information propagation for video understanding.
    Additionally, we use the semantically enhanced mask tokens and prompt embedding to finally generate the segmentation masks and class scores for generic segmentation, and instance embedding for video instance association.}
    \label{fig:model}
    \vspace{-3mm}
\end{figure*}

\textbf{Overall architecture.} The architecture of \name~is illustrated in Fig. \ref{fig:model}, which consists of a fine-grained pyramid visual encoder, a light-weight VLLM, and a segmentation predictor to generate segmentation masks, class scores, and instance embedding for video correspondence according to user’s instruction.
The proposed FVP module fuses multi-scale high-resolution visual features $f_{img}$ into a set of fine-grained tokens to ensure the injection of fine-grained visual information (Sec \ref{subsec:fvp}).
The VLLM takes three types of inputs: visual tokens encoded by the CLIP encoder, renewed fine-grained tokens, and prompt tokens for diverse instructions. The output embeddings of semantically enhanced mask tokens (Sec \ref{subsec:her}) and prompt tokens are further fed into the segmentation predictor for final segmentation results. Besides, we utilize the space-time information propagation and global prompt aggregation for comprehensive video understanding (Sec \ref{subsec:videoadapt}).
We train the LLM with LoRA for efficient parameter tuning.

\noindent \textbf{Visual Large Language Model}. 
We take a light-weight VLLM as our powerful multi-modal feature encoder, which contains a low-resolution vision encoder like CLIP~\cite{radford2021learning} and an efficient LLM.

Specifically, the model takes vision-prompt pairs $\{(\mathcal{V},\mathcal{P})\}$ as inputs, where $\mathcal{V}$ is resized to low resolution and then encoded by CLIP encoder $F_{CLIP}$ to get image features $f_{v}$. The $f_{v}$ is further projected and concatenated with other task-specific tokens to ensure the comprehensive understanding of multi-modal inputs through the fusion process of LLM $F_{LLM}$, where $G_{c}$ is the projection function and $E_{O}$ denotes the output embeddings of LLM. Formally,
\begin{equation}
     f_{v}=F_{CLIP}(\mathcal{V}), E_{O}=F_{LLM}(G_{c}(f_{v}), P, \mathcal{P}),
\label{Eq: PSALM}
\end{equation}
where $P$ denotes fine-grained tokens.
Furthermore, we manually extract semantic enhanced
mask tokens $E_\mathcal{Q}$ and prompt embedding $E_{\mathcal{P}}$
from $E_{O}$, which are further fed into the pre-trained segmentation predictor~\cite{cheng2022masked} to generate masks, class scores, and instance embedding for final segmentation results.


\noindent \textbf{Prompt design.} 
In order to accommodate the different segmentation tasks, we propose a flexible design for prompt $\mathcal{P}$. 
As illustrated above, we divide $\mathcal{P}$ into two formats: text prompts and visual prompts. 
To be specific, $\mathcal{P}$ contains the instructions $\mathcal{P_I}$ and task-specific conditions $\mathcal{P_C}$, where $\mathcal{P_I}$ instructs the model to perform different tasks while 
$\mathcal{P_C}$ indicates diverse conditions which are further used as classifiers to calculate the class scores of predicted masks.

For class-based segmentation tasks like panoptic segmentation, open-vocabulary segmentation (OVS), and video instance segmentation (VIS),  $\mathcal{P}$ can be demonstrated as $\mathcal{P_I}$: \emph{``Please segment all the positive objects according to the following potential categories.''} $\mathcal{P_C}$: \emph{``[category 1, category 2, category 3, ...]''}
\vspace{-0.5mm}

For referring and reasoning segmentation tasks like referring expression segmentation (RES), reasoning segmentation, referring video object segmentation (R-VOS), and ReasonVOS, $\mathcal{P}$ can be designed as
 $\mathcal{P_I}$: \emph{``Can you perform referring or reasoning segmentation according to the language expression?''}
 $\mathcal{P_C}$: \emph{``[referring / reasoning text]''}

For visual-guided segmentation tasks like interactive segmentation and video object segmentation (VOS), $\mathcal{P}$ can be designed as
 $\mathcal{P_I}$: \emph{``Please segment according to the given visual region reference''}
 $\mathcal{P_C}$: \emph{``[vision 1, vision 2, vision 3, ...]''}. 
Instead of using an additional region encoder to extract visual reference features~\cite{lin2023uninext}, we sample the CLIP visual features $f_{v}$ in VLLM according to the region coordinates and perform adaptive average pooling on them to form the final reference features for each visual prompt.

\noindent \textbf{Segmentation predictor.} Segmentation predictor $F_{p}$ generates the masks $m$, corresponding class scores $z$, and instance embedding $e$ through the similar process~\cite{cheng2022masked, gu2024dataseg} of three inputs: task-specific prompt embedding $\{E_{\mathcal{P}}^{k}\}_{k=1}^{K}$, the semantically enhanced mask tokens $\{E_{\mathcal{Q}}^{j}\}_{j=1}^{N}$ and the multi-scale visual features $f_{img}$, where $K$ and $N$ denote $K$ categories and $N$ mask proposals. Formally,
\begin{equation}
    \{m_{j}, z_{j}, e_{j}\}_{j=1}^{N} = F_{p}(\{E_{\mathcal{P}}^{k}\}_{k=1}^{K}, \{E_{\mathcal{Q}}^{j}\}_{j=1}^{N}, f_{img}),
\end{equation}
where $m_{j}\in \mathbb{R}^{H\times W}$ is the j-th mask proposal, $z_{j}\in \mathbb{R}^{K}$ denotes the class scores of $m_{j}$, and $e_{j}\in \mathbb{R}^{D}$ denotes the j-th instance embedding obtained from an extra embedding head only for video domain.
For video tasks, we adopt a frame-by-frame manner to get frame-level segmentation results for efficient training and inference processes.

\noindent \textbf{Training objectives.} The model can be trained jointly on multiple tasks using the unified loss $\mathcal{L}$. Specifically, we employ an autoregressive cross-entropy loss $\mathcal{L}_{text}$ for text prediction, a combination of per-pixel binary cross-entropy loss $\mathcal{L}_{bce}$ and DICE loss $\mathcal{L}_{dice}$ for mask supervision $\mathcal{L}_{mask}$, a cross-entropy loss $\mathcal{L}_{cls}$ for category classification, and a contrastive loss $\mathcal{L}_{ins}$ for instance association of video sequences following~\cite{IDOL}. $\lambda$ indicates their sum weight respectively. Formally,
\begin{equation}
\label{eq:total_loss}
\mathcal{L}= \mathcal{L}_{text}+\lambda_{mask}\mathcal{L}_{mask}+\lambda_{cls}\mathcal{L}_{cls}+\lambda_{ins}\mathcal{L}_{ins},
\end{equation}
\vspace{-6mm}
\begin{equation}
\label{eq:mask_loss}
\mathcal{L}_{mask}=\lambda_{bce}\mathcal{L}_{bce}+\lambda_{dice}\mathcal{L}_{dice},
\end{equation}

\begin{figure*}[t]
    \centering
    \includegraphics[width=\textwidth]{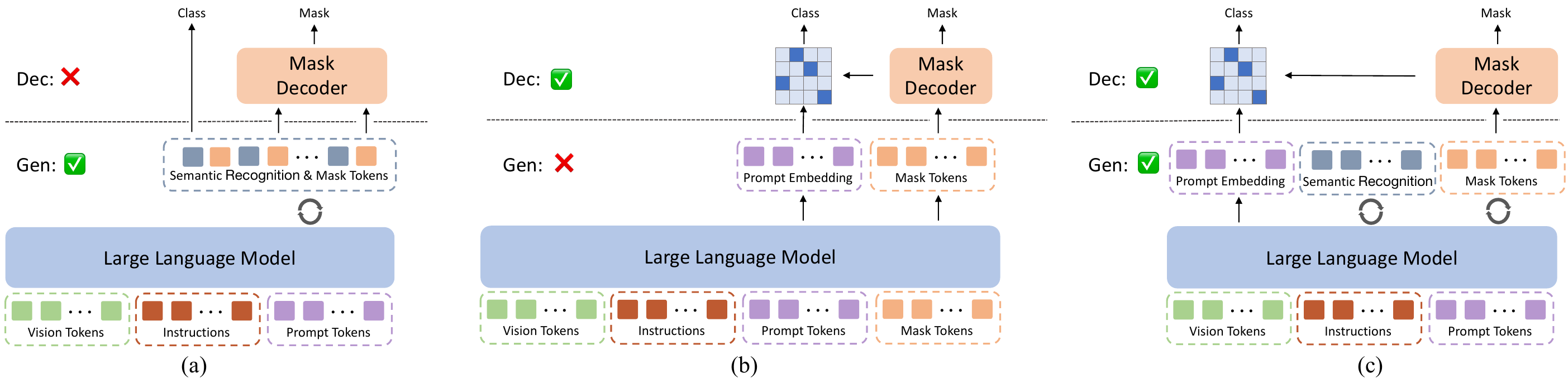}
     \vspace{-5mm}
    \caption{\textbf{The comparison of different recognition strategies}. (a) Generation-Only~\cite{Lai2023LISARS, Ren2023PixelLMPR}: both the semantic recognition (existing objects) and their mask tokens are generated by LLM. 
    (b) Decode-Only~\cite{zhang2024psalm,zhang2024omg}: prompt embedding and mask tokens are decoded from LLM. The present objects are then determined by their similarity scores. (c) Hybrid (ours): prompt embedding is decoded from LLM while the semantically enhanced mask tokens are generated by LLM. Their similarity scores reflect the objects' presence.
    }
    \label{fig:cmp}
    \vspace{-3mm}
\end{figure*}

\begin{figure}[t]
    \centering
    \includegraphics[width=0.48\textwidth]{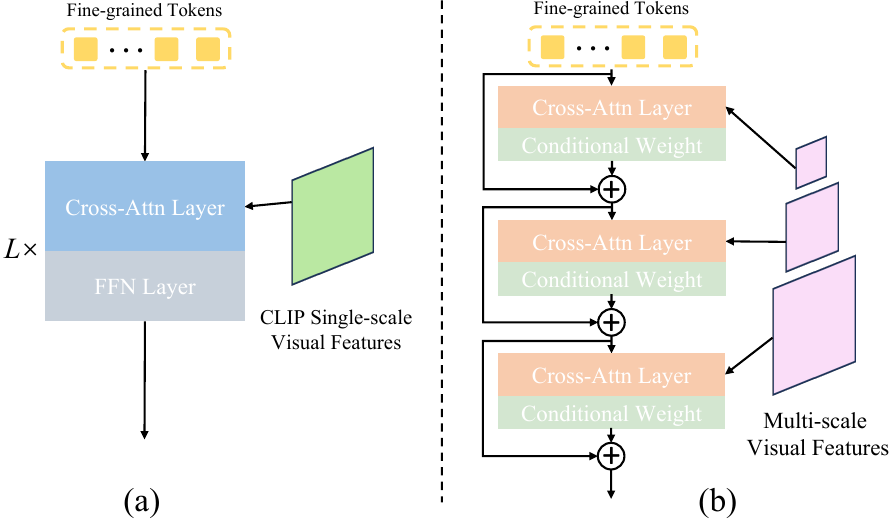}
    \caption{\textbf{Comparison between previous vision perceiver and our FVP}. (a): previous vision perceiver~\cite{li2023blip,bai2023qwen} uses the coarse single-scale CLIP visual features which are inadequate for fine-grained perception tasks.
    (b): FVP encodes the multi-scale visual features into fine-grained tokens.  }
    \label{fig:fvp}
    \vspace{-3mm}
\end{figure}

\noindent \textbf{Differences between \name~and previous methods.}
Previous universal segmentation methods~\cite{lin2023uninext,li2024omg,liu2024universal} lacking of VLLMs show inability in reasoning perception tasks
while our \name~demonstrates brilliant reasoning segmentation capability in complex scenarios.  
Besides, we make a significant generalization of the current VLLM-based segmentation methods~\cite{Lai2023LISARS,zhang2024psalm,yan2024visa,zhang2024omg} for more diverse segmentation tasks in both image and video domains using a single model framework.
Moreover, \name~differs from previous methods in the three designs elaborated in the following sections.

\subsection{Hybrid Entity Recognition}
\label{subsec:her}
As shown in Fig. \ref{fig:cmp} (a), predicting presented objects in the way of sequence generation (semantic prediction) tends to miss objects or produce repetitive predictions \cite{wei2024lasagna}. On the other hand, Fig. \ref{fig:cmp} (b), only using VLLM 
to embed class names (prompt tokens) as mask classifier at the decode stage disregards VLLM's powerful semantic recognition capability.
Consequently, we propose a hybrid approach that leverages LLM in both generation and decoding processes.

Instead of integrating mask tokens in input sequences and extracting the corresponding embedding from the one-pass forward output of VLLM, we instruct VLLM to generate the mask tokens preceded by the estimated objects' names. As illustrated in Fig. \ref{fig:cmp} (c),  VLLM is compelled to generate all the existing objects in the vision input and then the mask tokens. The semantically enhanced mask tokens contain valuable semantic integrated information about the image, which are subsequently used as input for the segmentation predictor to generate segmentation masks.

\begin{table*}[t]
  \centering
  \caption{ Comparison with the state-of-the-art models on the closed-set referring segmentation benchmarks (RefCOCO series) and more challenging generalized referring expression segmentation benchmark gRefCOCO. $\ddagger$ denotes models using pre-trained SAM~\cite{kirillov2023segment} for mask generation.
  * means using gRefCOCO for training while other methods are evaluated in zero-shot manners. Our \name~exhibits excellent performance over other zero-shot models like LaSagnA~\cite{wei2024lasagna} and PSALM~\cite{zhang2024psalm}.
  }
  \scalebox{0.84}{
    \begin{tabular}{c|l|ccc|ccc|cc|ccc}
    \toprule[1.1pt] 
    \multirow{2}{*}{ Type } & \multirow{2}{*}{ Method } & \multicolumn{3}{c|}{ RefCOCO } & \multicolumn{3}{c|}{ RefCOCO+ } & \multicolumn{2}{c|}{ RefCOCOg } & \multicolumn{3}{c}{ gRefCOCO }\\
    \cline { 3 - 13 } & & val & testA & testB & val & testA & testB & val(U) & test(U) & val & testA & testB\\
    \midrule[0.7pt]
    \multirow{4}{*}{\shortstack{Segmentation\\Specialist}} 
    & VLT~\cite{ding2021vision} & 67.5 & 70.5 & 65.2 & 56.3 & 61.0 & 50.1 & 55.0 & 57.7 & 52.5* & 62.2* & 50.5* \\
    & CRIS~\cite{wang2022cris} & 70.5 & 73.2 & 66.1 & 62.3 & 68.1 & 53.7 & 59.9 & 60.4 & 55.3* & 63.8* & 51.0* \\
    & LAVT~\cite{yang2022lavt} & 72.7 & 75.8 & 68.8 & 62.1 & 68.4 & 55.1 & 61.2 & 62.1 & 57.6* & 65.3* & 55.0* \\
    & PolyFormer-B~\cite{liu2023polyformer} & 74.8 & 76.6 & 71.1 & 67.6 & 72.9 & 59.3 &67.8 &69.1 & - & - & - \\
    \midrule[0.5pt] 
    \multirow{11}{*}{\shortstack{MLLM-based\\Segmentation Network}} 
    & LISA-7B~\cite{Lai2023LISARS} $\ddagger$ & 74.9 & 79.1 & 72.3 & 65.1 & 70.8 & 58.1 &  67.9 & 70.6 & 38.7* & 52.6* & 44.8* \\
    & PixelLM-7B~\cite{Ren2023PixelLMPR}& 73.0 &76.5 & 68.2 & 66.3 & 71.7 & 58.3 &69.3 & 70.5 & - & - & - \\
    & F-LMM-7B~\cite{wu2024f} $\ddagger$ & 76.1 & - & - & 66.4 & - & - & 70.1 & - & - & - & - \\ 
    & GSVA-7B~\cite{xia2023gsva} $\ddagger$ & 76.4 &77.4 &72.8 &64.5 &67.7 & 58.6 & 71.1 & 72.0 & 61.7* & 69.2* & 60.3* \\
    & GroundHog-7B~\cite{miao2023spectrum} & 78.5 & 79.9 & 75.7 & 70.5 & 75.0 & 64.9 & 74.1 & 74.6 & 66.7* & - & - \\
    & SAM4MLLM-7B~\cite{chen2025sam4mllm} $\ddagger$ & 79.6 & 82.8 & 76.1 & 73.5 & 77.8 & 65.8 & 74.5 & 75.6 & 66.3* & 70.1* & 63.2* \\
    & LaSagnA-7B~\cite{wei2024lasagna} $\ddagger$ & 76.8 & 78.7 & 73.8 & 66.4 & 70.6 & 60.1 & 70.6 & 71.9 & 38.1 & 50.4 & 42.1 \\
       
    & OMG-LLaVA ~\cite{zhang2024omg} & 78.0   & 80.3  & 74.1   & 69.1    & 73.1     & 63.0   & 72.9     & 72.9  & - & - & -  \\
    &GLaMM~\cite{rasheed2024glamm} $\ddagger$ & 79.5    & 83.2    & 76.9  & 72.6    & 78.7     & 64.6    &  74.2   & 74.9   & - & - & - \\
    & PSALM ~\cite{zhang2024psalm} & 83.6   & 84.7  & 81.6   & 72.9    & 75.5     & 70.1   & 73.8     & 74.4 & 42.0 & 52.4 & 50.6  \\ 
 
    & \textbf{\name} & \textbf{84.8} & \textbf{85.7} & \textbf{83.4} & \textbf{79.0} & \textbf{83.5} & \textbf{75.2} & \textbf{79.4} & \textbf{78.9} & \textbf{47.5} & \textbf{57.3} & \textbf{52.5} \\

    \bottomrule[1.1pt]
    \end{tabular}
}

\label{tab:ref}
\end{table*}

\begin{table*}[t]
  \centering
  \caption{ Comparison with the state-of-the-art models on more complex and challenging reasoning segmentation benchmarks: ReVOS in video domain and ReasonSeg in image domain. $\ddagger$ denotes the same meaning as Tab. \ref{tab:ref}.
  Our \name~outperforms all the previous VLLM-based models in both video and image reasoning segmentation tasks.
  }
  \scalebox{0.87}{
    \begin{tabular}{l|c|ccc|ccc|ccc|cc}
    \toprule[1.1pt] 
    \multirow{2}{*}{ Method } & \multirow{2}{*}{ Backbone } & \multicolumn{3}{c|}{ ReVOS-Reasoning } & \multicolumn{3}{c|}{ ReVOS-Referring } & \multicolumn{3}{c|}{ ReVOS-Overall } & \multicolumn{2}{c}{ ReasonSeg }\\
    \cline { 3 - 13 } & & $\mathcal{J}$ & $\mathcal{F}$ &  $\mathcal{J\&F}$ & $\mathcal{J}$ & $\mathcal{F}$ &  $\mathcal{J\&F}$ & $\mathcal{J}$ & $\mathcal{F}$ &  $\mathcal{J\&F}$ & gIoU & cIoU \\
    \midrule[0.7pt]

    LMPM~\cite{ding2023mevis} & Swin-T & 13.3 & 24.3 & 18.8 & 29.0 & 39.1 & 34.1 &  21.2 & 31.7 & 26.4 & - & - \\

    ReferFormer~\cite{wu2022language} & Video-Swin-B & 21.3 & 25.6 & 23.4 & 31.2 & 34.3 & 32.7 &  26.2 & 29.9 & 28.1 & - & - \\

     LISA-7B~\cite{Lai2023LISARS} $\ddagger$ & ViT-H & 33.8 & 38.4 & 36.1 & 44.3 & 47.1 & 45.7 &  39.1 & 42.7 & 40.9 & 52.9 & 54.0\\
    LaSagnA-7B~\cite{wei2024lasagna} $\ddagger$ & ViT-H & - & - & - & - & - & - & - & - & - & 48.8 & 47.2\\
    SAM4MLLM-7B~\cite{chen2025sam4mllm} $\ddagger$ & EfficientViT-SAM-XL1  & - & - & - & - & - & - & - & - & - & 46.7 & 48.1\\
    TrackGPT-13B~\cite{zhu2023tracking} $\ddagger$ & ViT-H & 38.1  & 42.9 & 40.5  & 48.3  & 50.6  & 49.5   & 43.2  & 46.8  & 45.0  & - & - \\
    VISA-7B ~\cite{yan2024visa} $\ddagger$ & ViT-H & 36.7  & 41.7 & 39.2  & 51.1  & 54.7  & 52.9   & 43.9  & 48.2  & 46.1  & 52.7 & \textbf{57.8} \\
    VISA-13B ~\cite{yan2024visa} $\ddagger$ & ViT-H & 38.3  & 43.5 & 40.9  & 52.3  & 55.8  & 54.1   & 45.3  & 49.7  & 47.5  & - & - \\

    \midrule[0.5pt] 

   
    \textbf{\name-3B} & Swin-B
 & \textbf{50.2} & \textbf{55.8} & \textbf{53.0} & \textbf{56.0} & \textbf{60.9} & \textbf{58.5} & \textbf{53.1} & \textbf{58.4}  & \textbf{55.7} & \textbf{59.2} & 56.7 \\

    \bottomrule[1.1pt]
    \end{tabular}
}

\label{tab:reason}
\end{table*}

\subsection{Fine-grained Visual Perceiver}
\label{subsec:fvp}
\textbf{Why twin-tower vision encoder}?
As shown in Fig. \ref{fig:fvp}, previous VLLMs and VLLM-based segmentation methods usually utilize the pre-trained CLIP encoder to obtain single-scale and low-resolution vision features interacted with diverse languages, which is insufficient for fine-grained image and video segmentation tasks.
Therefore, we adopt an extra pyramid vision encoder~\cite{cheng2022masked} to inject details-aware visual information. 

Specifically, we fuse multi-scale visual features into fine-grained tokens (stated as $P$ in Sec \ref{subsec:overview}) which can inject rich fine-grained visual information into the pre-trained VLLMs without excessive computation cost.
Formally, given the vision input $\mathcal{V}$, we leverage a pyramid vision encoder~\cite{cheng2022masked} $F_{seg}$ 
to get details-aware image features $f_{img}$. For the j-th scale and the previous fine-grained tokens $P_{j-1}$, the FVP module enriches each token through conditional weighted cross-attention: 

\begin{equation}
         \hat{P}_{j}=\textrm{MHCA}(P_{j-1}, G_{p}(f_{img}^{(j)})), 
    \label{eqn:gate1}
\end{equation}
\vspace{-5mm}
\begin{equation}
         P_{j}=P_{j-1} + \textrm{tanh}(\textrm{MLP}(\hat{P}_{j})) \cdot \hat{P}_{j},
    \label{eqn:gate2}
\end{equation}
where $\textrm{MHCA}$ denotes the Multi-Head Cross-Attention layer, $G_p$ is the projection function, $\textrm{tanh}$ is a normalization function and $\textrm{MLP}$ is a Multilayer Perceptron. The component of $\textrm{tanh}(\textrm{MLP}(\hat{P}_{j}))$ is the \emph{conditional weight} used to multiply the enriched fine-grained tokens $\hat{P}_{j}$ before the residual connection to the previous tokens $P_{j-1}$. Additionally, we initialize the weight value to zero to ensure the adaptation to diverse multi-scale image features while retaining the training stability.

\subsection{Temporal Adapter}
\label{subsec:videoadapt}
Video segmentation entails distinct challenges, requiring reasoning across multiple frames and the maintenance of temporal coherence. Existing VLLM-based methods exhibit limitations in addressing video perception tasks and lack specialized designs for comprehending temporal dynamics in video analysis.
To this end, we utilize global prompt aggregation and local space-time information injection in the time dimension to adapt to more complicated video perception tasks. 

\noindent \textbf{Global prompt aggregation}.
For the current prompt embedding $E_{\mathcal{P}}$ in the video object mask retrieval process, we leverage the adaptive average pooling strategy along the time dimension to aggregate global object and temporal information of previous $T$ frames. 
\begin{equation} \label{eq1}
E_{\mathcal{P}} =  AvgPool([E_{\mathcal{P}}^0, E_{\mathcal{P}}^1, ... , E_{\mathcal{P}}^T]),
\end{equation}

\noindent \textbf{Local space-time information injection}.
We propose a sequential renewal strategy for space-time information propagation based on fine-grained tokens $P$ to inject object information of adjacent frames. Formally,
\begin{equation} \label{eq2}
P_t =  G_{l}[F_{LLM}(P_{t-1})],
\end{equation}
where $P_t$ denotes the time-aware fine-grained tokens of the current $t$-th frame, $G_{l}$ is the projection function to transfer the previous features to the current space and align the feature dimensions.

The proposed global prompt aggregation and local space-time information injection within our temporal adapter facilitate the coalescence of both long-term and short-term vision-language information, which is essential for comprehensive video perception.

\begin{table*}[t]
  \centering
    \caption{ Quantitative results on the closed-set COCO-Panoptic segmentation, open-vocabulary segmentation (-OV) benchmarks. Our model \name~achieves remarkable performance compared with the previous state-of-the-art methods.}
\scalebox{0.81}{
  \begin{tabular}{c|l|c|cc|cc|c|c|c}
    \toprule[1.1pt]
    \multirow{2}{*}{Type} & \multirow{2}{*}{Method} & \multirow{2}{*}{Backbone} & \multicolumn{2}{c|}{COCO-Panoptic} & \multicolumn{2}{c|}{ADE-OV} & \multicolumn{1}{c|}{Citys-OV} & \multicolumn{1}{c|}{PC59-OV} & \multicolumn{1}{c}{PAS20-OV} \\
    \cline { 4 - 10 } & & & PQ & mIoU & PQ & mIoU & PQ & mIoU & mIoU  \\
    \midrule[0.7pt]
    \multirow{7}{*}
    {\shortstack{Segmentation \\ Specialist}}  &  Mask2former~\cite{cheng2022masked} & Swin-B  & 55.1 &  65.1 & - & - & - & - & - \\
    & OneFormer~\cite{jain2023oneformer} & Swin-L  & 57.9  & 67.4 & - & - & - & - & - \\
    & SEEM~\cite{seem}  &  DaViT-B & 56.1  & 66.3 & - & - & - & - & -  \\
    & MaskCLIP~\cite{maskclip} & ViT-L & 30.9 & 47.6 & 15.1 & \textbf{23.7} & - & 45.9 & - \\
    & DeOP~\cite{han2023open} & ResNet-101c & - & - & - & 22.9 & - & 48.8 & 91.7 \\


    & SimBaseline~\cite{xu2022simple} & ViT-B & - & - & - & 20.5 & - & 47.7 & 88.4 \\

    & DaTaSeg~\cite{gu2024dataseg}   &  ViTDet-B  & 52.8  & 62.7 & 12.3 & 18.3 & 28.0 & 51.1 & - \\

    \midrule[0.5pt]
    \multirow{3}{*}{\shortstack{MLLM-based\\Segmentation Network}}
    & OMG-LLaVA ~\cite{zhang2024omg} & ConvNeXt-L & 53.8 &  - & - & - & - & - & - \\
    & PSALM ~\cite{Lai2023LISARS} & Swin-B & 55.9 & 66.6 & 13.7 & 18.2 & 28.8 & 48.5 & 81.3 \\
    & \textbf{\name} & Swin-B & \textbf{61.2} & \textbf{77.2} & \textbf{16.1} & 22.3 & \textbf{31.1} & \textbf{64.6} & \textbf{92.1} \\
    \bottomrule[1.1pt]
  \end{tabular}
}
  \label{tab:cocoseg}
\end{table*}

\begin{table}[t]
  \centering
  \caption{ Results of common video segmentation benchmarks, including DAVIS17, Ref-YouTube-VOS, Ref-DAVIS17, and YouTube-VIS 2019. $\ddagger$ denotes the same meaning as Tab. \ref{tab:ref}.
  }
  \scalebox{0.64}{
    \begin{tabular}{l|c|c|c|c|c}
    \toprule[1.1pt] 
    \multirow{2}{*}{ Method } & 
    \multirow{2}{*}{ Backbone} & 
    \multicolumn{1}{c|}{  DAVIS17 } &
    \multicolumn{1}{c|}{  Ref-YT } & \multicolumn{1}{c|}{ Ref-DAVIS }  & \multicolumn{1}{c}{ YT-VIS } \\
    \cline { 3 - 6 } & & $\mathcal{J\&F}$ & $\mathcal{J\&F}$ & $\mathcal{J\&F}$ & mAP  \\
    \midrule[0.7pt]
    SEEM~\cite{seem} & DaViT-B & 62.8 & - & - & -   \\
    
    OMG-Seg~\cite{li2024omg}  & ConvNeXt-L & 74.3 & - & - & 56.4  \\

    ReferFormer~\cite{wu2022language}  & Video-Swin-B & - & 62.9 & 61.1 & -  \\
    OnlineRefer~\cite{wu2023onlinerefer}  & Swin-L & - & 63.5 & 64.8 & -  \\
    UNINEXT~\cite{lin2023uninext}  & ConvNeXt-L & 77.2 & 66.2 & 66.7 & \textbf{64.3}  \\

    \midrule[0.5pt]

    LISA-7B~\cite{Lai2023LISARS} $\ddagger$ & ViT-H & - & 53.9 & 64.8 & -  \\
    VISA-13B~\cite{yan2024visa} $\ddagger$ & ViT-H & - & 63.0 & 70.4 & -  \\
    VideoLISA-3.8B~\cite{bai2024one} $\ddagger$ & ViT-H & - & 63.7 & 68.8 & -  \\
    
     \textbf{\name-3B} & Swin-B & \textbf{77.6} & \textbf{68.5} & \textbf{71.2} & 53.8 \\
    
    \bottomrule[1.1pt]
    \end{tabular}
    }

\vspace{-3mm}
  
\label{tab:exp-video}
\end{table}

\section{Experiments}
\textbf{Datasets}.
We use the one-stage training strategy to train \name~with the multi-dataset and multi-task manners. For image segmentation, we use COCO Panoptic~\cite{Lin2014MicrosoftCC}, RefCOCO series~\cite{yu2016modeling,nagaraja2016modeling}, COCO-Interactive, and ReasonSeg~\cite{Lai2023LISARS}.
For video segmentation, we utilize DAVIS-2017 datasets~\cite{caelles20182018}, Ref-Youtube-VOS~\cite{seo2020urvos}, YouTube-VIS 2019~\cite{yang2019video}, and ReVOS~\cite{yan2024visa}. Besides, we use LLAVA-150k~\cite{liu2024visual} to maintain the vision-language conversation capability of VLLM (we show the results on Multi-modal benchmarks in the supplementary material).

\noindent \textbf{Implementation details}
We load the pre-trained weights of Mipha~\cite{zhu2024comprehensive} for our VLLM, and Maks2Former~\cite{cheng2022masked} for our segmentation predictor. We use three layers of FVP for fine-grained information fusion and utilize LoRA~\cite{hu2021lora} to finetune the LLM efficiently. We train \name~on all the tasks jointly for 160k iterations using a batch size of 32 on 8 NVIDIA A100 GPUs, which means each task takes approximately 16k iterations. We employ the AdamW optimizer with a learning rate of $4\times10^{-5}$ and with a cosine schedule. All the hyper-parameters in the loss $\mathcal{L}$are assigned values 1.0.

\begin{table*}[t]
  \centering
  \caption{
  The mutual influence between different tasks. Task-specific means training task-specific models only on data from corresponding tasks, Refer+Reason denotes the model is trained on referring and reasoning segmentation data, and Video and Image denote different training visual types: training on video data and image data, respectively.
  }
  \scalebox{0.76}{
    \begin{tabular}{c|c|c|c|ccc|cc|ccc|c}
    \toprule[1.1pt] 
    \multirow{2}{*}{Task-specific} & 
    \multirow{2}{*}{ Refer+Reason } &
    \multirow{2}{*}{ Video } &
    \multirow{2}{*}{ Image } &
    \multicolumn{3}{c|}{  RefCOCO } &
    \multicolumn{2}{c|}{  COCO } & \multicolumn{3}{c|}{ ReVOS }  & \multicolumn{1}{c}{ YT-VIS } \\
    \cline { 5 - 13 } & & & & val & testA & testB & PQ & mIoU & Reasoning & Referring & Overall & mAP  \\
    \midrule[0.7pt]
    \Checkmark &  &  &  & 83.8 & 85.9 & 82.2 & 60.8 & 75.1 & 51.2 & 56.6 & 53.9 & 50.7  \\
     & \Checkmark &  &  & 83.3 & 84.9 & 80.9 & - & - & 53.1 & 57.3 & 55.2 & -  \\
     &  &  & \Checkmark & 85.6 & 86.1 & 82.4 & 60.9 & 76.5 & - & - & - & -  \\
     &  & \Checkmark &  & - & - & - & - & - & 51.1 & 57.0 & 54.1 & 50.4  \\
    
     & \Checkmark & \Checkmark & \Checkmark & 84.8 & 85.7 & 83.4 & 61.2 & 77.2 & 53.0 & 58.5 & 55.7 & 53.8 \\ 
    
    \bottomrule[1.1pt]
    \end{tabular}
    }
  
\label{tab:ab-data}
\end{table*}

\begin{table*}[t]
  \centering
  \caption{ The comparison of different LLMs and backbone usages. w/o CLIP means without using CLIP vision encoder. 
  }
  \scalebox{0.8}{
    \begin{tabular}{c|c|cc|ccc|c|c|c}
    \toprule[1.1pt]  
    \multirow{2}{*}{ Method} & 
    \multirow{2}{*}{ LLM } & 
    \multicolumn{2}{c|}{  COCO } & \multicolumn{3}{c|}{ ReVOS } &
    \multicolumn{1}{c|}{  ADE-OV } &
    \multicolumn{1}{c|}{  PC59-OV  } &
    \multicolumn{1}{c}{  PAS20-OV  } \\
    \cline { 3 - 10 }  & & PQ & mIoU & Reasoning & Referring & Overall & mIoU & mIoU & mIoU \\
    \midrule[0.7pt]

    LISA~\cite{Lai2023LISARS} & Vicuna-7B & - & - & 36.1 & 45.7 & 40.9 & - & - & - \\
     VISA ~\cite{yan2024visa} & Vicuna-13B & - & - & 40.9 & 54.1 & 47.5 & - & - & - \\
    PSALM(w/o CLIP) ~\cite{Lai2023LISARS} & Phi-1.5-1.3B & 55.9 & 66.6 & - & - & - & 18.2 & 48.5 & 81.3 \\

    \midrule[0.5pt]

    \name~(w/o CLIP) & Phi-1.5-1.3B & 61.1 & 76.0 & 44.0 & 49.7 & 46.9 & 18.9 & 60.0 & 90.6 \\
    \name & Phi-1.5-1.3B & 60.9 & 76.7 & 50.8 & 57.0 & 53.9 & 20.3 &61.5 & 90.8 \\
    \name & Phi-2-2.7B & \textbf{61.2} & \textbf{77.2} & \textbf{53.0} &  \textbf{58.5} & \textbf{55.7} & \textbf{22.3} & \textbf{64.6} & \textbf{92.1} \\

    \bottomrule[1.1pt]
    \end{tabular}
    }
    
\vspace{-2mm}

\label{tab:ab-llms}
\end{table*}

\begin{table}[t]
  \centering
  \caption{ Ablation on the core components of \name. 
  FVP and HER denote the proposed Fine-grained Visual Perceiver and Hybrid Entity Recognition modules.
  }
  \scalebox{0.9}{
    \begin{tabular}{c|c|c|cc|c}
    \toprule[1.1pt] 
    \multirow{2}{*}{ FVP} & 
    \multirow{2}{*}{ HER } & 
    \multicolumn{1}{c|}{  YT-VIS } &
    \multicolumn{2}{c|}{  COCO } &  \multicolumn{1}{c}{ RefCOCO } \\
    \cline { 3 - 6 }  & & mAP & PQ & mIoU & cIoU  \\
    \midrule[0.7pt]

    &  & 48.4 & 54.8 & 66.2  & 82.8  \\
     \Checkmark &  & 50.8 & 55.8 & 66.6  &  84.6 \\
     & \Checkmark & 52.0 & 59.7 & 74.6 & 84.3  \\ 
     \Checkmark & \Checkmark & \textbf{53.8} & \textbf{61.2} & \textbf{77.2} & \textbf{84.8} \\

    \bottomrule[1.1pt]
    \end{tabular}
    }
  
\label{tab:ab-component}
\end{table}

\subsection{Comparisons with State-of-the-Arts}


\noindent\textbf{Referring expression segmentation results}.
We compare \name~with the state-of-the-art methods on the benchmarks RefCOCO/+/g~\cite{yu2016modeling,nagaraja2016modeling} and more challenging generalized referring expression segmentation benchmark gRefCOCO~\cite{liu2023gres}. in Tab.~\ref{tab:ref}. Based on the versatile and adaptable design of \name, our model achieves state-of-the-art performance on all the referring datasets. Specifically, \name~surpasses the current SOTA by a large margin, reaching 79.7 cIoU on RefCOCO+ val (+6.8 over PSALM).
Besides, Our model shows superiority in challenging G-RES tasks compared with previous \textbf{zero-shot} methods, demonstrating the robustness and generalization ability of~\name.

\noindent\textbf{Reasoning segmentation results}.
We compare \name~with the state-of-the-art methods on image reasoning segmentation (ReasonSeg~\cite{Lai2023LISARS}) and reasoning video object segmentation (ReVOS~\cite{yan2024visa}) in Tab.~\ref{tab:reason}.
Our \name~achieves superior performance on reasoning tasks, significantly surpassing previous state-of-the-art methods (+12.1 on ReVOS-Reasoning), which shows \name~powerful reasoning capability of tackling complex scenarios.



\noindent\textbf{Generic image segmentation results}.
We show the performance of \name~on  COCO-Panoptic~\cite{Lin2014MicrosoftCC} and open-vocabulary segmentation~\cite{zhou2019semantic,Cordts2016Cityscapes,mottaghi2014role,everingham2010pascal} tasks in Tab.~\ref{tab:cocoseg}.
\name~achieves excellent performance compared with both specialist models and VLLMs-based methods on both closed-set and open-vocabulary segmentation tasks. Specifically, \name~surpasses the VLLM-based PSALM by a significant margin (+5.3 on COCO PQ, and +10.6 on mIoU), which demonstrates our powerful capabilities of handling complex semantic perception and segmentation tasks.
Besides, we show the results of COCO-Interactive in the supplementary material.

\noindent\textbf{Common video segmentation results}.
We compare \name~with previous video segmentation methods in Tab.~\ref{tab:exp-video}, including visual-prompted semi-supervised VOS (DAVIS17 val), text-prompted referring video object segmentation (Ref-YouTube-VOS, Ref-DAVIS17) and video instance segmentation (YouTube-VIS 2019).
\name~shows promising results over previous unified segmentation methods~\cite{li2024omg,lin2023uninext}.
Besides, \name~ performs more video perception tasks than previous VLLM-based models~\cite{yan2024visa,bai2024one}.



\subsection{Ablations}

\noindent\textbf{The mutual influence between different tasks.} 
Our model can be trained and inferred across multiple tasks and datasets simultaneously.
We evaluate the mutual impact of different tasks in Tab.~\ref{tab:ab-data}. The results show that joint training can enhance the model performance compared with the task-specific model.
Besides, the performance of video segmentation tasks can be improved significantly by adding the image training datasets.
This demonstrates the generalization and self-consistency of our \name~to perform universal segmentation.

\noindent\textbf{Effect of different LLMs and vision backbone.} In Tab.~\ref{tab:ab-llms}, we evaluate the effect of different sizes of LLMs and vision backbone.
Our \name~achieves excellent performance using smaller LLMs and vision encoder compared with the previous SOTA models like VISA\cite{yan2024visa} and PSALM\cite{zhang2024psalm}.
Besides, the performance of \name~can be further improved by using the more powerful LLM (Phi-2-2.7B~\cite{javaheripi2023phi}).

\begin{table}[t]
  \centering
  \caption{ Ablation on the Fine-grained Visual Perceiver design. CW denotes the Conditional Weight illustrated in Sec.~\ref{subsec:fvp}, and Scale denotes the total scale in the proposed FVP module.
  }
  \scalebox{0.84}{
    \begin{tabular}{c|c|c|cc|c}
    \toprule[1.1pt] 
    \multirow{2}{*}{ CW } & 
    \multirow{2}{*}{ Scale } & 
    \multicolumn{1}{c|}{  YT-VIS } &
    \multicolumn{2}{c|}{  COCO } & \multicolumn{1}{c}{ RefCOCO } \\
    \cline { 3 - 6 } & & mAP & PQ & mIoU & cIoU  \\
    \midrule[0.7pt]

    & single-layer & 49.7 & 55.8 & 68.0  &  83.7 \\
    & multi-layers & 50.4  & 58.9 & 73.4  & 84.5 \\

    \Checkmark & multi-layers & \textbf{53.8} & \textbf{61.2} & \textbf{77.2} & \textbf{84.8} \\

    \bottomrule[1.1pt]
    \end{tabular}
    }
\vspace{-3mm}
  
\label{tab:ab-fvp}
\end{table}

\noindent\textbf{Ablation on the proposed components.} We assess the effectiveness of our proposed FVP module and Hybrid Entity Recognition strategy.
As shown in Tab.~\ref{tab:ab-component}, with our fine-grained visual integration and hybrid entity semantic enhancement, the segmentation accuracy can be enhanced significantly (+5.4 on YT-VIS, +6.4 on COCO panoptic PQ).

\noindent\textbf{Design of the Fine-grained Visual Perceiver}. In the FVP module, we combine multi-scale visual features into fixed perception queries using the condition-wise cross-attention layers to extract rich visual details from different scales of the pyramid encoder. 
As shown in Tab.~\ref{tab:ab-fvp}, together with the conditional weight and the multi-scale design, our model makes a significant improvement on both image and video segmentation tasks.

\noindent\textbf{Effect of temporal adapter.} We evaluate the effectiveness of the proposed temporal adapter including global prompt aggregation (global) and local space-time information injection (local) in Tab.~\ref{tab:ab-temporal}.
Incorporating both global and local components, the temporal adapter significantly enhances model performance across multiple video segmentation tasks.

\begin{table}[t]
  \centering
  \caption{ Ablation on the temporal adapter for video tasks, including global prompt aggregation (global) and local space-time information injection (local).
  }
  \scalebox{0.9}{
    \begin{tabular}{c|c|c|c|c}
    \toprule[1.1pt] 
    \multirow{2}{*}{ Global } & 
    \multirow{2}{*}{ Local } & 
    \multicolumn{1}{c|}{  Ref-DAVIS17 } &
   \multicolumn{1}{c|}{ ReVOS }  & \multicolumn{1}{c}{ YT-VIS } \\
    \cline { 3 - 5 } & & $\mathcal{J\&F}$  & $\mathcal{J\&F}$ & mAP  \\
    \midrule[0.7pt]
      &  & 67.3 & 54.1 & 47.9 \\
    \Checkmark &  & 68.8 & 54.5 & 48.5  \\
     & \Checkmark & 69.3 & 54.8 & 50.2  \\
    \Checkmark & \Checkmark & \textbf{71.2} & \textbf{55.7} & \textbf{53.8}  \\
    
    \bottomrule[1.1pt]
    \end{tabular}
    }

\vspace{-3mm}
  
\label{tab:ab-temporal}
\end{table}

\section{Conclusion}
In this study, we aim to present \name, the first VLLM-based universal segmentation model designed for pixel-level image and video perception, encompassing a wide range of generic segmentation and complex reasoning tasks.
We propose the Hybrid Entity Recognition and Fine-grained Visual Perceiver to leverage the recognition capacity of VLLMs more effectively and enhances the VLLM's ability by capturing diverse levels of visual information without incurring excessive computational costs. With additional Temporal Adapter, \name~can tackle challenging video tasks by incorporating global and local information. \name~surpasses existing methods on complex reasoning segmentation and traditional perception tasks. The insights presented in this work expand the possibilities of VLLMs in visual perception and lay a foundation for future research on the integration of vision-language models.

{
    \small
    \bibliographystyle{ieeenat_fullname}
    \bibliography{arxiv}
}

\clearpage
\setcounter{page}{1}
\maketitlesupplementary
\appendix

\section{Additional Implementation Details}
\label{sec:implement_details}

\subsection{Evaluation Metrics}

In our experiments, we use the widely used metrics to evaluate the performance of our \name~on various segmentation tasks consistent with previous studies. Specifically, cumulative Intersection-over-Union (cIoU) for referring expression segmentation (RES), interactive segmentation, and generalized referring expression segmentation (G-RES), cIoU and the average of all per-image Intersection-over-Unions (gIoU) for reasoning segmentation task, region similarity $\mathcal{J}$ and contour accuracy $\mathcal{F}$ for reasoning video object segmentation (ReasonVOS), video object segmentation (VOS), referring video object segmentation (R-VOS), panoptic quality (PQ), mean intersection-over-Union (mIoU) for image generic segmentation, and mean average precision (mAP) for video instance segmentation (VIS).

\subsection{Training Details}
In our experiments, we use Phi-2~\cite{javaheripi2023phi} with 2.7B parameters as our Large Language Model,  SigLIP~\cite{zhai2023sigmoid} as our vanilla encoder, and Swin-B~\cite{liu2021swin} as our pyramid encoder.
We use PyTorch to implement our \name~and use Deepspeed zero-1 optimization for efficient training. 
Furthermore, the vanilla encoder and pyramid encoder are kept frozen, the LLM is finetuned with LORA (rank=8), the FVP, HER, and segmentation predictor are fully trained. Our codes and model weights will be publicly released.

\section{Additional Experimental Results}

\subsection{Multi-modal Question Answering Benchmarks}
Our \name~is the first VLLM-based universal segmentation model for pixel-level image and video perception with complex reasoning and conversation capabilities, which is capable of tackling vision-language comprehension tasks. 
Therefore, we evaluate our model on various Multi-modal question answering benchmarks. 
As shown in Tab.~\ref{tab:mllm_results}, our \name~achieves comparable performance compared with previous VLLMs like 
InstructBLIP~\cite{instructblip}, Qwen-VL~\cite{bai2023qwen}, and LLaVA-1.5~\cite{liu2024visual} with fewer model parameters, demonstrating the insights into the model’s powerful conversational and reasoning capabilities.

\begin{table}[ht]
    \centering
    \caption{Quantitative results of our \name~on Multi-modal question answering benchmarks. \name~achieves promising performance compared with previous VLLMs in several widely used Multi-modal benchmarks.}
    \scalebox{0.7}{
    \begin{tabular}{l|c|c|c|c|c|c}
    \toprule[1.1pt]  
    Method & 
    LLM & 
    MMB  & VQA\textsuperscript{v2}  &  GQA & POPE & SQA  \\
    \midrule[0.5pt]

    BLIP-2~\cite{li2023blip} & Vicuna-13B & - & 65.0 & 41.0 & 85.3 & 61.0 \\
    InstructBLIP~\cite{instructblip} & Vicuna-7B & 36.0 & - & 49.2 & - & 60.5  \\
    InstructBLIP~\cite{instructblip} & Vicuna-13B & - & - & 49.5 & 78.9 & 63.1  \\
    Shikra~\cite{chen2023shikra} & Vicuna-13B & 58.8 & 77.4 & - & - & -  \\
    Qwen-VL~\cite{bai2023qwen} & Qwen-7B & 38.2 & 78.8 & 59.3 & - & 67.1  \\
    Qwen-VL-Chat~\cite{bai2023qwen} & Qwen-7B & 60.6 & 78.2 & 57.5 & - & 68.2  \\
    LLaVA-1.5~\cite{liu2024visual} & Vicuna-7B & 64.3 & 78.5 & 62.0 & 85.9 & 66.8  \\

    \name & Phi-2-2.7B & 67.9 & 78.2 & 60.9 & 86.6 & 66.2 \\ 
    
    \bottomrule[1.1pt]
    \end{tabular}
    }
    
    \label{tab:mllm_results}
\end{table}

\subsection{Interactive Segmentation}

We also evaluate \name~on the COCO-Interactive validation set for the interactive segmentation task. As shown in Tab.~\ref{tab:interactive_seg}, our \name~achieves promising performance on various visual prompt types. 
Notably, our model surpasses previous segmentation specialists such as SAM~\cite{kirillov2023segment}, which utilizes a larger vision backbone and much more high-quality training data, and SEEM~\cite{seem}. 
However, the VLLM-based model PSALM~\cite{zhang2024psalm} exhibits superior performance in the interactive segmentation task. 
We hypothesize that this discrepancy arises from differences in feature scale utilization during the visual prompt sampling process: PSALM~\cite{zhang2024psalm} employs the visual prompt features derived from a high-resolution Swin-based vision encoder, whereas \name~utilizes features from a more streamlined CLIP-based visual encoder.

\begin{table}[ht]
  \centering
    \caption{ Quantitative results on COCO-Interactive benchmark.}
\scalebox{0.81}{
  \begin{tabular}{l|c|c|c|c|c}
    \toprule[1.1pt]
    Method & Backbone & Box & Scribble & Mask & Point \\
    \midrule[0.7pt]
    
    SAM~\cite{kirillov2023segment} & ViT-B & 
    68.7 & - & - & 33.6 \\
    SAM~\cite{kirillov2023segment} & ViT-L & 
    71.6 & - & - & 37.7 \\
    SEEM~\cite{seem}  &  DaViT-B & 
    42.1 & 44.0 & 65.0 & 57.8 \\
    
    PSALM ~\cite{Lai2023LISARS} & Swin-B & 
    80.9 & 80.0 & 82.4 & 74.0 \\
    \name & Swin-B & 
    77.3 & 75.2 & 79.5 & 63.4 \\
    \bottomrule[1.1pt]
  \end{tabular}
}
  \label{tab:interactive_seg}
\end{table}


\begin{table*}[ht]
  \centering
  \caption{ The comparison of different settings between our model and previous segmentation specialists and VLLM-based segmentation methods. Generic Seg denotes common class-based segmentation, such as panoptic segmentation and semantic segmentation. Open-set denotes the open-vocabulary segmentation. \name~can perform more comprehensive segmentation tasks in one model. 
  }
  \scalebox{0.65}{
    \begin{tabular}{c|l|c|cc|ccccc}
    \toprule[1.1pt] 
    \multirow{2}{*}{ Type } & \multirow{2}{*}{ Method } & \multirow{2}{*}{ Multi-task Training } & \multicolumn{2}{c|}{ Visual Type } & \multicolumn{5}{c}{ Task Type } \\
     & & & Image-level & Video-level & Referring Seg & Reasoning Seg  & Generic Seg & Interactive Seg & Open-set\\
    \midrule[0.7pt]
    \multirow{10}{*}{\shortstack{Segmentation\\Specialist}}
    & Mask2former~\cite{cheng2022masked} &  & \Checkmark &  & & & \Checkmark & &  \\
    & OneFormer~\cite{jain2023oneformer} &  & \Checkmark &  & & & \Checkmark & &  \\
    & VLT~\cite{ding2021vision} &  & \Checkmark &  & \Checkmark &   \\
    & LAVT~\cite{yang2022lavt} &  & \Checkmark &  & \Checkmark &   \\
    & PolyFormer~\cite{liu2023polyformer} &  & \Checkmark &  & \Checkmark &   \\
    & ReferFormer~\cite{wu2022language} &  &  & \Checkmark & \Checkmark &   \\
    & OnlineRefer~\cite{wu2023onlinerefer} &  &  & \Checkmark & \Checkmark &   \\
    & SEEM~\cite{seem} & \Checkmark & \Checkmark  & \Checkmark & \Checkmark & & \Checkmark  & \Checkmark & \Checkmark  \\
    & UNINEXT~\cite{lin2023uninext}  & \Checkmark & \Checkmark  & \Checkmark & \Checkmark & & \Checkmark  & \Checkmark & \Checkmark  \\
    & OMG-Seg~\cite{li2024omg}& \Checkmark & \Checkmark  & \Checkmark & & & \Checkmark  & \Checkmark & \Checkmark  \\
    
    \midrule[0.5pt] 
    
    \multirow{7}{*}{\shortstack{VLLM-based\\Segmentation Network}} 
    & LISA~\cite{Lai2023LISARS} & \Checkmark & \Checkmark &  & \Checkmark & \Checkmark \\
    & PixelLM~\cite{Ren2023PixelLMPR} & \Checkmark & \Checkmark &  & \Checkmark & \Checkmark \\
    & GSVA~\cite{xia2023gsva} & \Checkmark & \Checkmark &  & \Checkmark &  \\
    & LaSagnA~\cite{wei2024lasagna} & \Checkmark & \Checkmark &  & \Checkmark & \Checkmark & \Checkmark \\
    & OMG-LLaVA ~\cite{zhang2024omg} & \Checkmark & \Checkmark &  & \Checkmark & & \Checkmark \\
    & PSALM ~\cite{zhang2024psalm} & \Checkmark & \Checkmark & \Checkmark & \Checkmark & & \Checkmark &  \Checkmark &  \Checkmark  \\
     & VISA ~\cite{yan2024visa} & \Checkmark & \Checkmark & \Checkmark &  \Checkmark &  \Checkmark \\
    
    & \name~(Ours) & \Checkmark & \Checkmark & \Checkmark &  \Checkmark &  \Checkmark & \Checkmark &  \Checkmark &  \Checkmark \\

    \bottomrule[1.1pt]
    \end{tabular}
}

\label{tab:task-compare}
\end{table*}

\section{Comparison of different settings}
\label{sec:comparison_setting}
We also make setting comparisons between different models and our \name.  
As shown in Tab.~\ref{tab:task-compare}, \name~can handle more comprehensive segmentation tasks than previous segmentation specialists and MLLM-based methods. Firstly, \name~can tackle both image-level and video-level perception tasks in one model enjoying the benefits of multi-task joint training.
Secondly, \name~performs various segmentation tasks, including long-text prompted referring and reasoning segmentation, category prompted generic segmentation, visual prompted interactive segmentation, and open-vocabulary segmentation.

\section{Qualitative Results}
In this section, we present more qualitative results to better demonstrate the segmentation capabilities of our \name~involving various tasks in image and video domains.

\subsection{Referring Expression Segmentation (RES)}
\begin{figure*}[t]
    \centering
    \includegraphics[width=\textwidth]{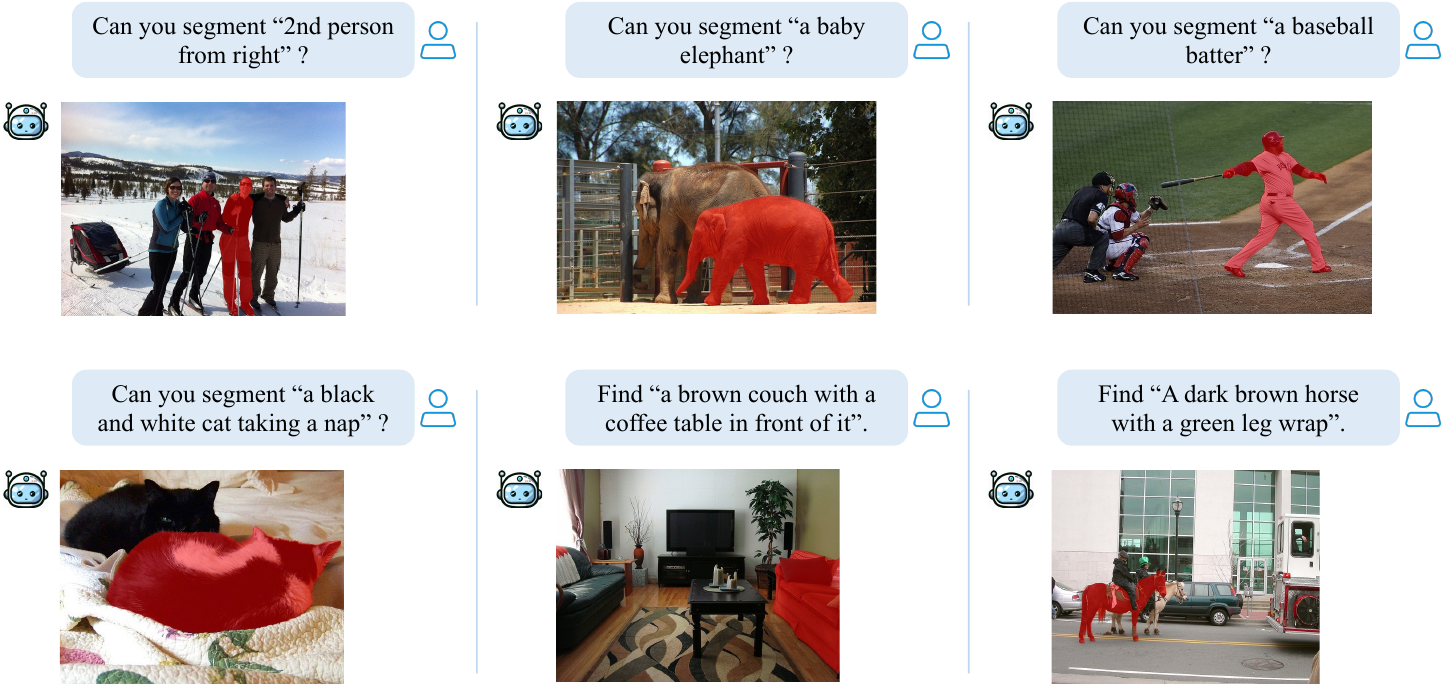}
    \caption{Qualitative results of \name’s capability in referring expression segmentation. 
    }
    \label{fig:visualize_res}
\end{figure*}

Fig. \ref{fig:visualize_res} shows the visualization of \name~on referring segmentation benchmarks (RefCOCO/+/g).
Our model can effectively grasp the true meaning conveyed by the referring text and provide accurate pixel-level segmentation masks.

\subsection{Interactive Segmentation}
\begin{figure*}[t]
    \centering
    \includegraphics[width=\textwidth]{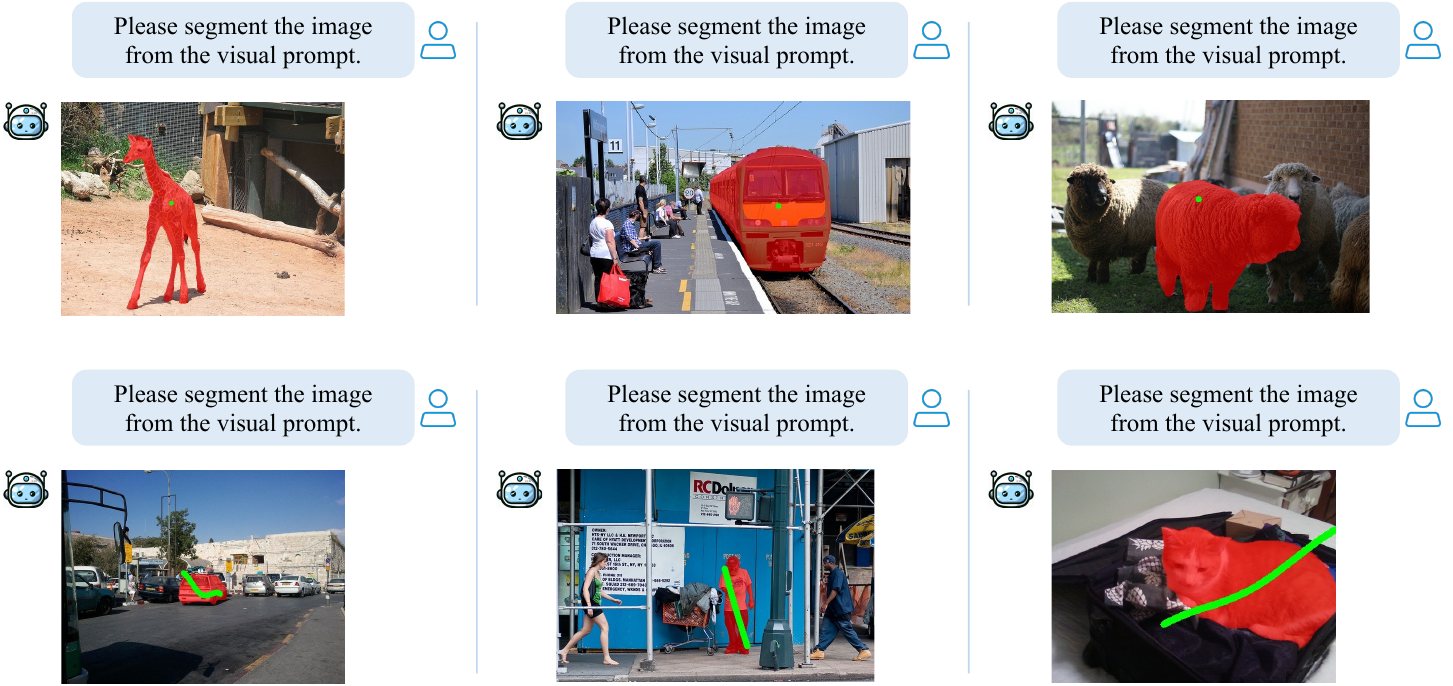}
    \caption{Qualitative results of \name~in interactive segmentation. The green marker indicates the provided visual prompts, such as point and scribble.
    }
    \label{fig:visualize_interact}
\end{figure*}

Fig. \ref{fig:visualize_interact} presents the effectiveness of our \name~in understanding the visual prompt and outputting the corresponding segmentation masks for the interactive segmentation tasks.

\subsection{Panoptic Segmentation}
\begin{figure*}[t]
    \centering
    \includegraphics[width=\textwidth]{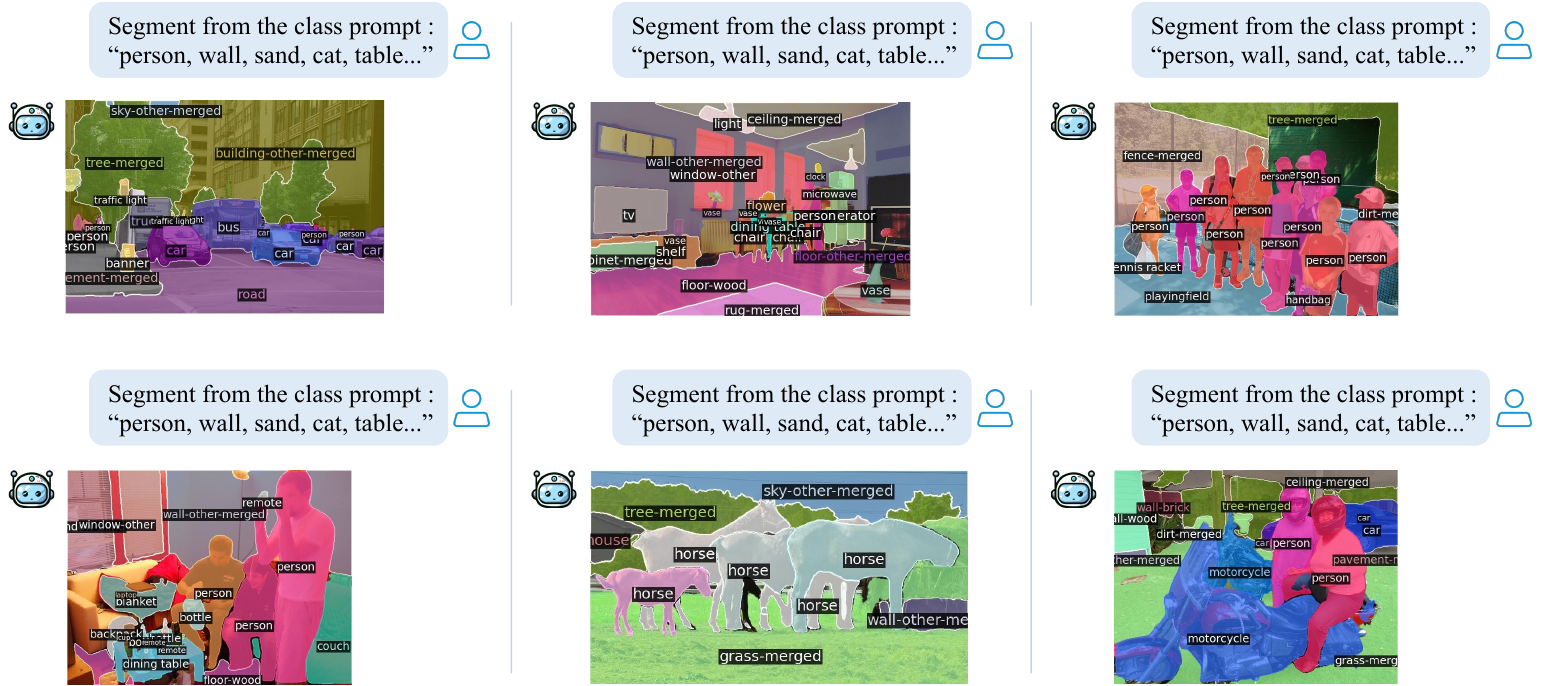}
    \caption{Qualitative results of \name~in panoptic segmentation.
    }
    \label{fig:visualize_pano}
\end{figure*}

Fig. \ref{fig:visualize_pano} shows the qualitative results of \name~in panoptic segmentation tasks, which needs
both semantic and instance level dense predictions.

\subsection{Reasoning Segmentation}
\begin{figure*}[t]
    \centering
    \includegraphics[width=\textwidth]{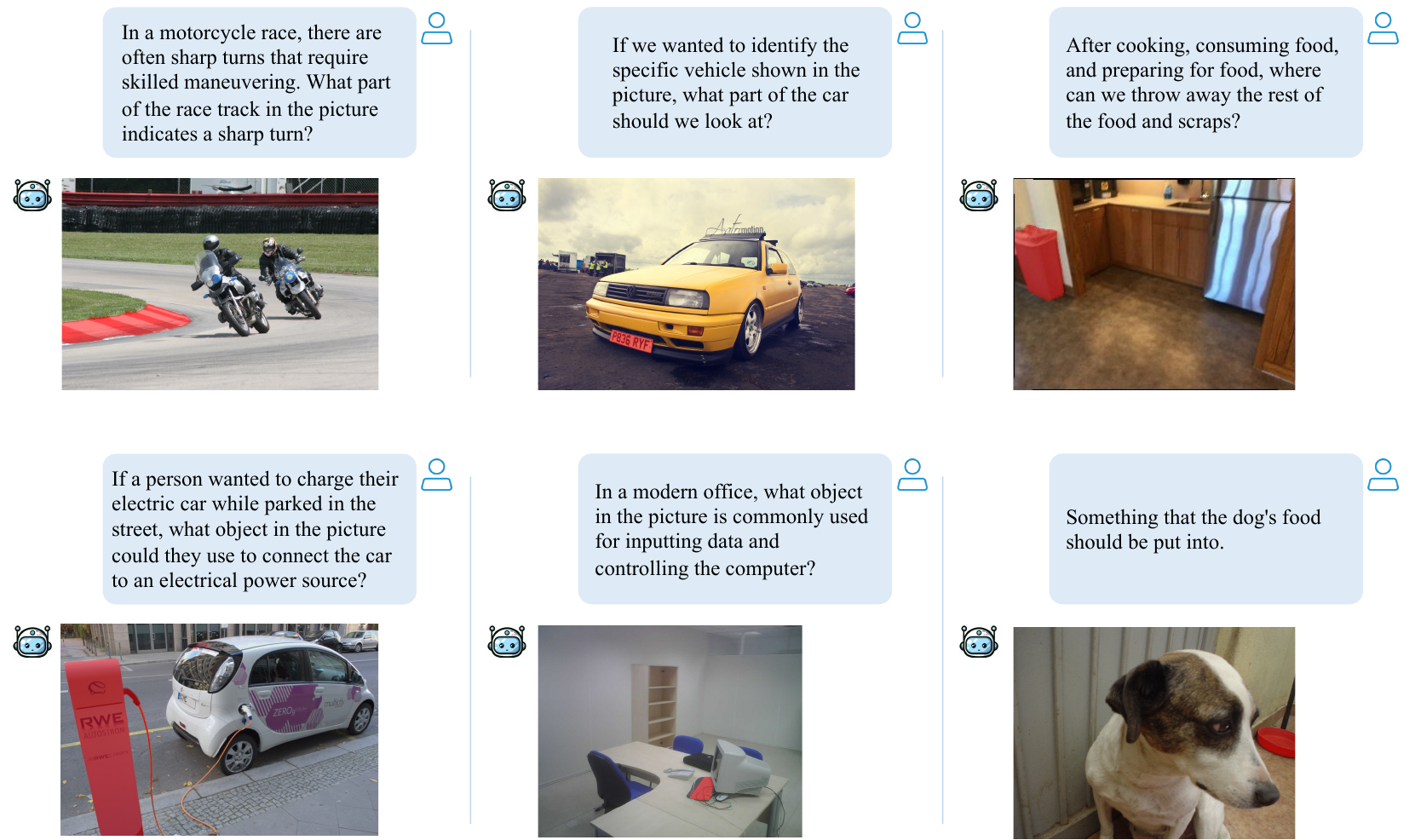}
    \caption{Qualitative results of \name~in 
    reasoning segmentation.
    }
    \label{fig:visualize_reason}
\end{figure*}

Fig. \ref{fig:visualize_reason} presents the effectiveness of our \name~in understanding the complex question and perform segmentation according to the reasoning process. 

\subsection{Reasoning Video Object Segmentation (ReasonVOS)}
\begin{figure*}[t]
    \centering
    \includegraphics[width=\textwidth]{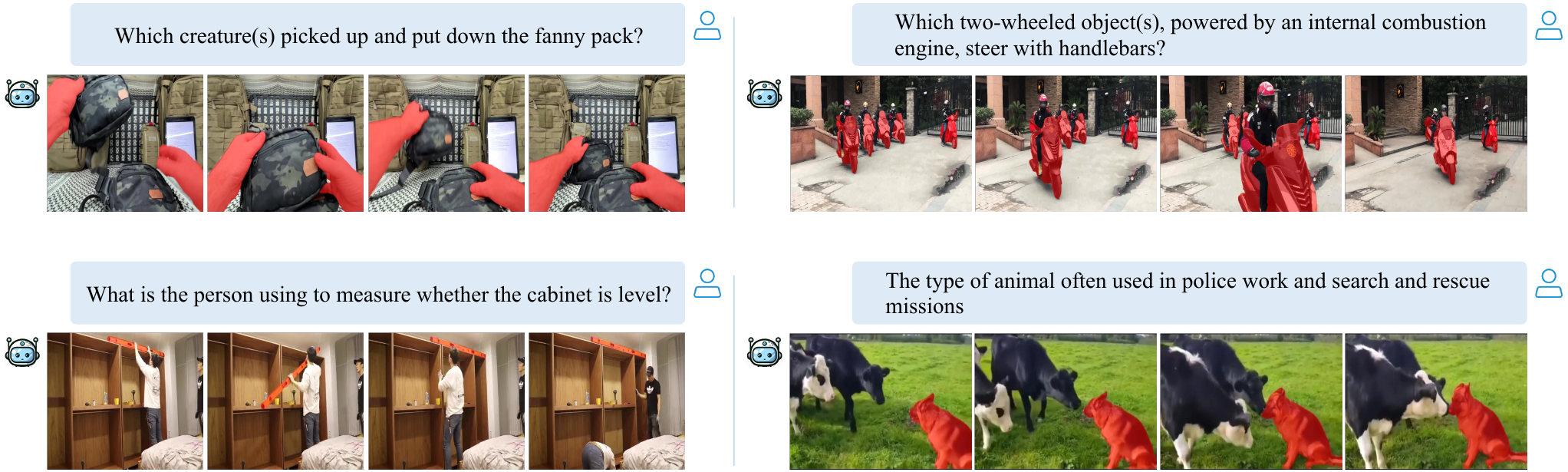}
    \caption{Qualitative results of \name~demonstrate its capability in the complex reasoning video object segmentation task, effectively managing challenging video data and producing temporally consistent results following the reasoning process.
    }
    \label{fig:visualize_revos}
\end{figure*}

Fig. \ref{fig:visualize_revos} shows the effectiveness of \name~in comprehending both the reasoning questions and temporal coherence. \name~is capable of producing segmentation masks that maintain consistency across temporal sequences.

\subsection{Video Object Segmentation (VOS)}
\begin{figure*}[t]
    \centering
    \includegraphics[width=\textwidth]{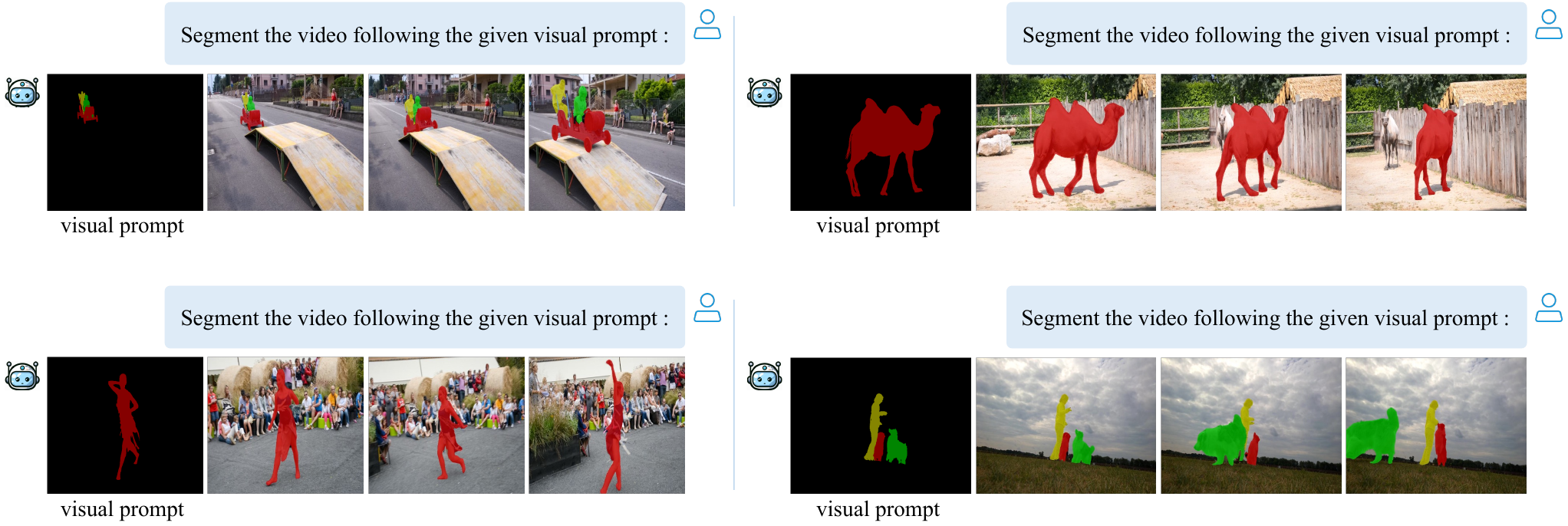}
    \caption{Qualitative results of \name~in semi-supervised video object segmentation tasks. With the visual prompts provided by the ground truth object masks of the first frame, \name~demonstrates its ability to achieve accurate segmentation while maintaining temporal consistency.
    }
    \label{fig:visualize_vos}
\end{figure*}

The qualitative results of our method, \name, are illustrated in Fig. \ref{fig:visualize_vos}, demonstrating its capability in interpreting the visual prompt, provided by the ground truth object masks of the first frame, and producing accurate segmentation masks that maintain temporal consistency.

\subsection{Video Instance Segmentation (VIS)}
\begin{figure*}[t]
    \centering
    \includegraphics[width=\textwidth]{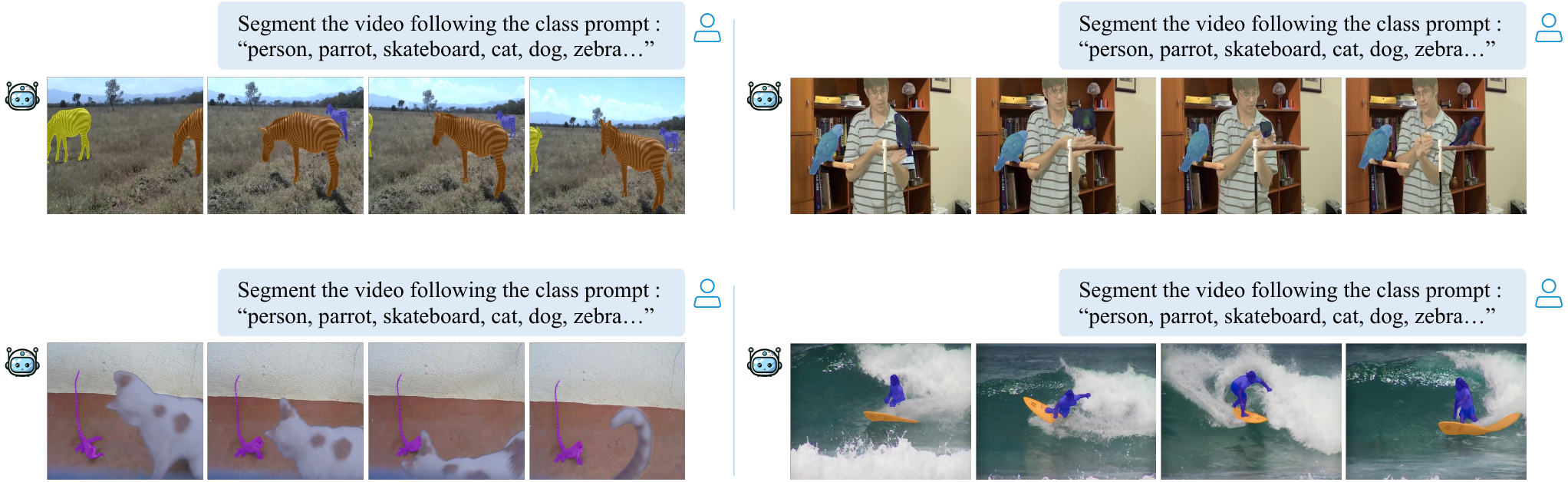}
    \caption{Qualitative results of \name~in video instance segmentation tasks. Utilizing the class text prompts and instance tracking strategies, \name~exhibits its capability to achieve precise segmentation while ensuring temporal consistency.
    }
    \label{fig:visualize_vis}
\end{figure*}

Fig. \ref{fig:visualize_vis} illustrates the effectiveness of \name~in performing instance-level video segmentation with class prompts, and executing accurate segmentation with instance tracking throughout the entire video.

\end{document}